\documentclass[letterpaper]{article} 
\usepackage{aaai24}  
\usepackage{times}  
\usepackage{helvet}  
\usepackage{courier}  
\usepackage[hyphens]{url}  
\usepackage{graphicx} 
\usepackage{natbib}  
\usepackage{caption} 
\usepackage{algorithm}
\usepackage{algorithmic}

\usepackage{booktabs}
\usepackage{multirow}

\usepackage{newfloat}
\usepackage{listings}

\usepackage{amssymb}
\usepackage{amsmath}
\usepackage{subcaption}
\usepackage{graphicx}
\usepackage{makecell}
\usepackage{color}

\DeclareCaptionStyle{ruled}{labelfont=normalfont,labelsep=colon,strut=off} 
\floatstyle{ruled}
\newfloat{listing}{tb}{lst}{}
\floatname{listing}{Listing}
\title{Graph Context Transformation Learning for Progressive Correspondence Pruning}
\author {
    Junwen Guo\textsuperscript{\rm 1,2},
    Guobao Xiao\textsuperscript{\rm 1}\thanks{Corresponding author},
    Shiping Wang\textsuperscript{\rm 2},
    Jun Yu\textsuperscript{\rm 3}
}
\affiliations {
    \textsuperscript{\rm 1}School of Electronics and Information Engineering, Tongji University,\\
    \textsuperscript{\rm 2}College of Computer and Data Science, Fuzhou University, \\
    \textsuperscript{\rm 3}School of Computer Science and Technology, Hangzhou Dianzi University\\
    mxyttkx@163.com,
    x-gb@163.com,
    shipingwangphd@163.com,
    yujun@hdu.edu.cn
}

\usepackage{bibentry}

\begin{document}

\title{Graph Context Transformation Learning for Progressive Correspondence Pruning}

\maketitle

\begin{abstract}
  Most of existing correspondence pruning methods only concentrate on gathering the context information as much as possible while neglecting effective ways to utilize such information. In order to tackle this dilemma, in this paper we propose Graph Context Transformation Network (GCT-Net) enhancing context information to conduct consensus guidance for progressive correspondence pruning. Specifically, we design the Graph Context Enhance Transformer which first generates the graph network and then transforms it into multi-branch graph contexts. Moreover, it employs self-attention and cross-attention to magnify characteristics of each graph context for emphasizing the unique as well as shared essential information. To further apply the recalibrated graph contexts to the global domain, we propose the Graph Context Guidance Transformer. This module adopts a confident-based sampling strategy to temporarily screen high-confidence vertices for guiding accurate classification by searching global consensus between screened vertices and remaining ones. The extensive experimental results on outlier removal and relative pose estimation clearly demonstrate the superior performance of GCT-Net compared to state-of-the-art methods across outdoor and indoor datasets. The source code will be available at: https://github.com/guobaoxiao/GCT-Net/.
\end{abstract}

\section{Introduction}
\par
\begin{figure}[t]
    \centering
    \begin{subfigure}{0.49\linewidth}
        \includegraphics[width=1\linewidth]{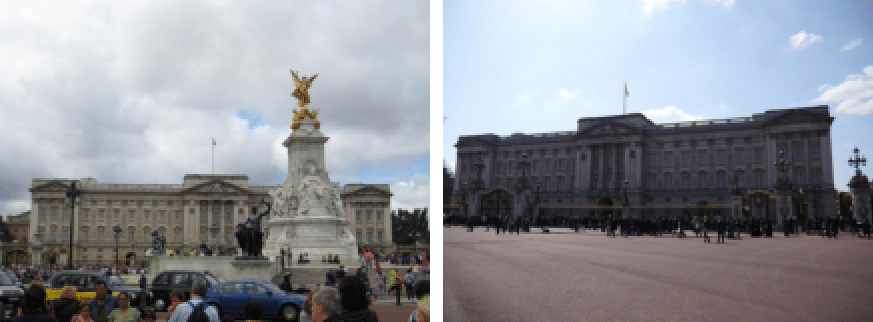}
        \caption{Original Image}
        \label{fig:subfig1a}
    \end{subfigure}
    \begin{subfigure}{0.49\linewidth}
        \includegraphics[width=1\linewidth]{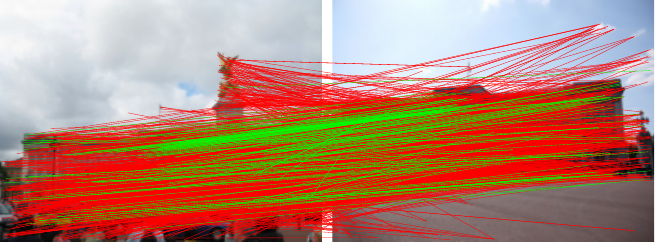}
        \caption{Initial Correspondence Set}
        \label{fig:subfig1b}
    \end{subfigure}

    \begin{subfigure}{0.49\linewidth}
        \includegraphics[width=1\linewidth]{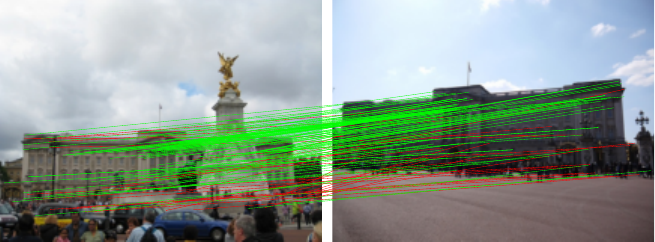}
        \caption{Graph Context Enhancement}
        \label{fig:subfig2a}
    \end{subfigure}
    \begin{subfigure}{0.49\linewidth}
        \includegraphics[width=1\linewidth]{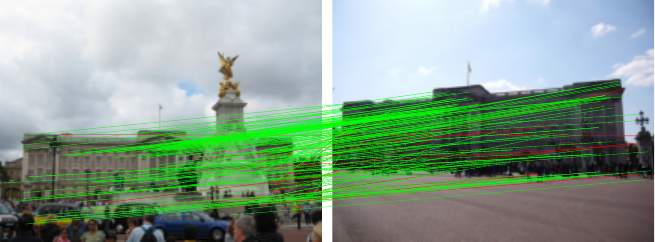}
        \caption{Global Consensus Guidance}
        \label{fig:subfig2b}
    \end{subfigure}
    \caption{Visualization results of the procedure for correspondence pruning using GCT-Net, with the red and green lines representing incorrect and accurate correspondences, respectively.}
    \label{fig:fm}
\end{figure}

 Two-view correspondence pruning methods strive to form robust correspondences between two sets of interest points to lay the foundation for many computer vision tasks, such as, Structure from Motion (SfM) \cite{sarlin2023pixel}, Simultaneous Localization and Mapping (SLAM) \cite{johari2023eslam} and Image Fusion \cite{tang2023rethinking}. Correspondence pruning involves three steps: keypoints and relating descriptors extraction, the initial correspondence set establishment and outlier ($i.e.$ false correspondence) removal. More specifically, we employ established methods, such as SuperPoint \cite{detone2018superpoint} and SIFT \cite{lowe2004distinctive}, to generate keypoints and computer their descriptors at first. Subsequently, the initial correspondence set is generated by applying the nearest matching algorithm to the descriptors. However, the initial correspondence set often contains numerous outliers (shown in Fig.~\ref{fig:subfig1b}) due to the limitations of local descriptor representation and the presence of low-quality images. Therefore, the third step, identifying and eliminating outliers, is indispensable. As illustrated in Fig.~\ref{fig:subfig2a} and Fig.~\ref{fig:subfig2b}, outliers are removed and most of inliers are preserved, enhancing the available insights for post-processing endeavors.

\par Correspondence pruning methods are mainly evolved into two distinct factions, $i.e.$, tradition methods and learning-based methods. RANSAC \cite{fischler1981random} and its modifications \cite{barath2019magsac,chum2005matching,chum2005two,torr2000mlesac} are representative of traditional methods. These methods adopt a sampling-verification loop to retain most of correspondences adhering to a specific geometric model. However, in scenarios with a high proportion of outliers, the runtime of these methods will rapidly increase and simultaneously, the quality of the results will significantly deteriorate. In our task, it is common for the initial correspondence set to have an outlier proportion exceeding $80\%$, rendering these methods inapplicable.

\par The most of learning-based advancements approach the correspondence pruning as a binary classification problem and achieve remarkable potential. But this treatment also poses formidable challenges: ($1$) unordered data should be handled appropriately to ensure its permutation invariance. ($2$) local context and global context should be adequately mined to provide the basis to identify outliers.

\par For example, LFGC-Net \cite{yi2018learning} and OANet \cite{zhang2019learning} employ the PointNet-like architecture \cite{qi2017pointnet} which embeds Context Normalization (CN) into Multi-Layer Perceptrons (MLPs) to deal with each correspondence individually for gaining the global context. CLNet \cite{zhao2021progressive}, MS$^{2}$DG-Net \cite{dai2022ms2dg} and NCM-Net \cite{liu2023progressive} all construct the graph network via K-Nearest Neighbor (KNN) to build relationships among adjacent correspondences for searching and aggregating local context. All these methods aim to obtain adequate local and global context information while preserving the permutation invariance of input data. However, their primary emphasis lies in acquiring abundant context information, overlooking the practical utilization of such context knowledge.
\par In this paper, we propose the Graph Context Enhance Transformer (GCET) block that not only gathers multi-branch graph contexts but also thoroughly mines and emphasizes respective and common significant context information via self-attention and cross-attention to boost the discriminating capability of the network. In specific, we first employ KNN to construct the graph network where each node represents a correspondence and each edge denotes a relationship between two correspondences. Next, we transform the graph network filled with a wealth of context information into two completely different types of graph context to receive various evidence. In one type, local context is aggregated by MLPs and maxpooling, which, although sacrificing a considerable amount of structure information, guarantees the reliability of graph context. In the other type of graph context, local context is gathered by affinity-based convolution which captures the vast majority of relationships in the graph structure while preserving contaminated information. It is feasible to immediately integrate the complementary graph contexts from multiple branches, but it is not the optimal choice because there is untapped potential yet to be explored. Here, we utilize self-attention to amplify the crucial parts and reliable dependencies among correspondences for emphasizing the distinctiveness of graph context within each branch. In parallel, we employ cross-attention to uncover and enhance the shared importance between different salient graph contexts. Finally, the discriminative fusion strategy is applied to absorb highlight components of graph contexts and discard redundant portions meanwhile.

\par Moreover, in order to further apply the fused graph context to the global domain, we present the Graph Context Guidance Transformer (GCGT) block which adopts the score-based sampling to select a set of candidates and utilize transformer to guide spatial consensus at the global level through sampled candidates. Specifically, we first employ the linear layer to score the confidence values for each correspondence. Then, we select a set of correspondences with high scores as the global guiding source and regard original correspondences as the guiding target. It is worth noting that before the guiding procedure, we also perform the cluster operation on the guiding source and target to enhance their reliability and reduce computational overhead. At last, the guiding source and target are fed into transformer to steer correspondences with high global consistency and shape long-range dependencies. Additionally, we also capture short-range spatial dependencies to complement the output of transformer.

\par Our contributions are three-fold: ($1$) We propose the GCET block which generates multi-branch graph contexts with respective characteristics and employs self-attention and cross-attention to emphasize individual nature and shared crucial knowledge for better absorbing advantages from each branch. ($2$) Drawing on the foundation of global consensus and guided by the principle of distinct inliers to guide hidden inliers, we design the GCGT block which samples credible inliers to direct remaining inliers exhibiting high spatial consensus. ($3$) By combining the GCET block and GCGT block, we develop an effective Graph Context Transformation Network (GCT-Net) for outlier removal and relative pose estimation, achieving the state-of-the-art performance on both outdoor and indoor datasets.

\begin{figure*}[t]
\centering
\centerline{\includegraphics[width=1.05 \textwidth, height=0.195\textwidth]{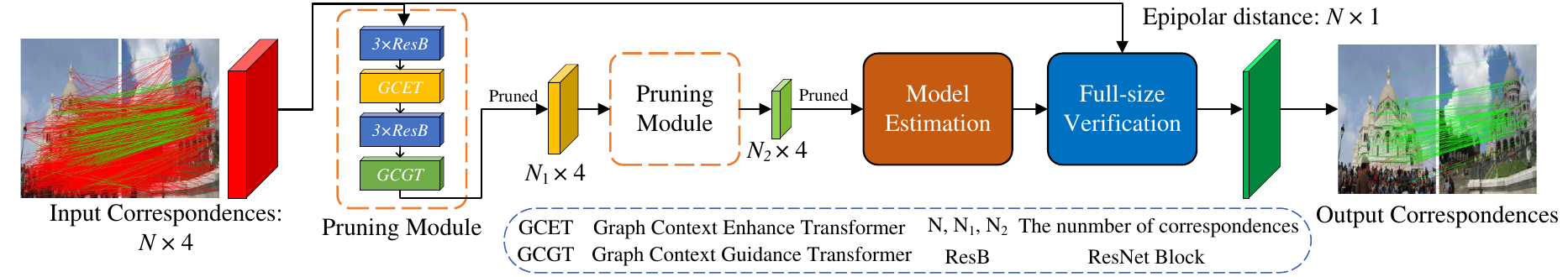}}
\caption{The detailed pipeline of our proposed GCT-Net. Feeding the initial correspondences: $N\times4$, GCT-Net progressively aggregates context information and filtrates outliers in the pruning module to reduce the negative impact from polluted information. Next, the pruned correspondences are delivered to Model Estimation to calculate the essential matrix. Finally, by performing Full-size Verification with generated matrix, the network outputs epipolar distances to determine whether correspondences are inliers or outliers in the end.}
\label{fig:GCT}
\end{figure*}

\section{Related Work}
\label{sec:Relatedwork}
\subsection{Learning-Based Correspondence Pruning Methods}
\par The advent of deep learning has provided many new inspirations for tackling outlier rejection. As a pioneer in this field, LFGC-Net \cite{yi2018learning}, driven by \cite{qi2017pointnet,qi2017pointnet++}, subdivides the correspondence pruning task into outlier/inlier labeling and essential matrix regression. Besides, it further designs a permutation-equivariant architecture, which integrates CN into MLPs, thereby dealing with unordered data while obtaining the global context. The most of subsequent works adopt the de facto framework and incorporate or modify components to gain context information to enhance the network performance. For example, in order to overcome the disturbance of contaminated information caused by CN, ACNet \cite{sun2020acne} transforms CN into attentive ones to discriminately treat various information. OANet \cite{zhang2019learning} clusters correspondences by the differentiable pooling layer and recovers the original order of correspondences based on the differentiable unpooling layer for exploiting potential local context as well as reducing computation overhead. CLNet \cite{zhao2021progressive} first constructs the local graph for each correspondence to gather local context and then connects all local graphs to generate a global one to acquire abundant global context. MS$^2$DG-Net \cite{dai2022ms2dg} leverages the combination of maxpooling and self-attention to progressively update the local graphs for multi-level context. Diverging from the aforementioned approaches, which generates a solitary graph context and directly employs it in a superficial manner, our method generates multi-branch graph contexts with respective characteristics and knowledge and refine graph contexts through both self-interactions and collaborative interactions.

\subsection{Attention Mechanism in Correspondence Pruning}
\par Currently, attention mechanisms have been extensively employed in many computer tasks, including semantic segmentation \cite{kirillov2023segment}, image fusion \cite{li2023feature} and so on. For the field of correspondence pruning, the introduction of attention mechanisms is beneficial to focus on inlier information and suppress redundant information, but it still necessitates some appropriate modifications. For instance, SENet \cite{hu2018squeeze} is a simple yet efficient channel attention mechanism that emphasizes important knowledge in the channel dimension via the squeeze-and-excitation (SE) block. However, it prioritizes the global aspect and neglects the demand for local aspect in correspondence pruning. Therefore, MSA \cite{zheng2022msa} introduces the multi-scale attention by remoulding the SE block, carrying out information recalibration from multiple perspectives for accurate inlier/outlier classification. Additionally, CA \cite{hou2021coordinate} and CBAM \cite{woo2018cbam}, integrating the spatial attention mechanisms, further consider the weight allocation in the spatial dimension. But their effectiveness is limited for correspondence pruning methods. The emergence of vanilla Transformer brings new prospects for addressing the problem of capturing long-range spatial dependencies, but simultaneously introduces various challenges. Firstly, considering that the number of correspondence $N$ typically falls within the range of $1500$ to $2000$ in correspondence pruning, the computation complexity of Transformers, which is $O(N^{2}\cdot D)$, results in a substantial computational burden. Although some varieties of transformer, like ViT \cite{dosovitskiy2020image} and Swin-Transformer \cite{liu2021swin}, reduce the computational load, the patch strategy can negatively impact the handling of unordered data. Secondly, in the initial correspondence set, outliers constitute the majority, and their presence can interfere with similarity computation, significantly diminishing the reliability of the output attention map. Therefore, we propose the GCGT block which mitigates the impact of contaminated information during interaction process and reduces computational overhead in a scaling approach. This transformation renders Transformer-like architecture well-suited for correspondence pruning.

\section{Methodology}
\label{sec:network}
\subsection{Problem Formulation}
\par Given a putative set of image pairs $(I,I^{'})$, we first employ off-the-shelf keypoint extraction methods \cite{lowe2004distinctive,detone2018superpoint} to search interest points and calculate concerning descriptors. Afterwards, the initial correspondence set ${Q}=[{q}_1, {q}_2,{q}_3,\dots, {q}_N] \in R^{N\times4}$ consisted of $N$ correspondences is generated by roughly matching keypoints through descriptor similarities. As the basic element in ${Q}$, ${q}_i$ represents the $i$-th correspondence which involves two coordinates normalized by camera intrinsics in respective image. However, the brute-force matching process contributes to an overwhelming proportion of outliers in the initial correspondence set. Consequently, an efficient correspondence pruning method should be developed to conduct more accurate correspondence classification and relative pose estimation.
\par In pursuit of this objective, we propose Graph Context Transformation Network (GCT-Net) and illustrate its network architecture in Fig.~\ref{fig:GCT}. We adopt the progressive pruning strategy \cite{zhao2021progressive} in our network which is capable of gradually screening outliers and thus mitigates the negative impact of contaminated information. With the perspective of pruning module, the input data passes through $3$ ResNet blocks, the Graph Context Enhance Transformer (GCET), another $3$ ResNet blocks and Graph Context Guidance Transformer (GCGT). ResNet blocks are used to boost representation ability of the network, GCET aims to enhance the converted graph context, and GCGT further leverages enhanced context to guide remaining inliers. In general, we first utilize two series-connected pruning modules to deal with the input correspondence set $Q$. The operations in these two modules can be respectively expressed as: $(Q_{1},o_{1})={f_{\sigma1}(Q)}$ and $(Q_{2},o_{2})={f_{\sigma2}(Q_{1})}$ where $\sigma1$ and $\sigma2$ are their related parameters. Here, $Q_{1}\in\mathbb{R}^{N_{1}\times4}$ and $Q_{2}\in\mathbb{R}^{N_{2}\times4}$ denote the pruned correspondence set ($N>N_{1}>N_{2}$) and $o_{1}\in\mathbb{R}^{N_{1}\times1}$ and $o_{2}\in\mathbb{R}^{N_{2}\times1}$ represent the output logit values. Based on logit values, we sort correspondences in descending order and preserve $50\%$ of correspondences by pruning the lower $50\%$.. It is worth noting that the feature map of $Q_{2}$ is preserved and additionally passed through a linear layer to estimate the inlier weight set $w$. The next step ($i.e.$ model estimation), regarding $w$ as supplementary input, involves and $Q_{2}$ with $w$ to execute the parametric model calculation ($i.e.$, estimate essential matrix $\hat{E})$. Finally, we leverage $\hat{E}$ combined with $Q$ to carry out the full-size verification which can retrieve the falsely removed inliers during sequential pruning process. In short, the model estimation and the full-size verification can be expressed as:
\begin{align}
\hat{E}&=H(Q_{2},w),\\
ED&=V(\hat{E},Q),
\end{align}
where $H(\cdot,\cdot)$ denotes the weighted eight-point algorithm \cite{ma2021image} and $V(\cdot,\cdot)$ represents the full-size verification operation to measure the epipolar distance set of all correspondences $(i.e., {ED}=[{ed}_1, {ed}_2,\dots, {ed}_N] \in R^{N\times1})$. Each correspondence ${q}_i$ corresponds to a polar distance ${ed}_i$, and we classify ${q}_i$ into inliers when ${ed}_i$ is less than an artificially set threshold.
\subsection{Graph Context Enhance Transformer}
\begin{figure}[ht]
\centering
\centerline{\includegraphics[width=0.37\textwidth]{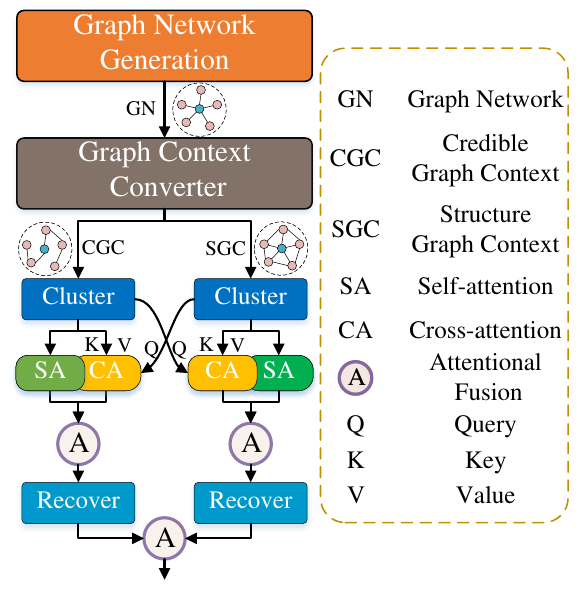}}
\caption{Structure diagram of our proposed graph context enhance transformer.}
\label{fig:GCET}
\end{figure}
\par Collecting abundant local context is highly beneficial for accurate correspondence pruning. The graph network plays a significant role in establishing and exploring relationships among neighbors. \cite{zhao2021progressive} and \cite{dai2022ms2dg} leverage the nature of graph network to generate graph contexts with respective advantages. However, the converted graph contexts are not thoroughly explored and refined resulting in a lost opportunity to substantially improve the effectiveness of subsequent tasks.
\par As shown in Fig.~\ref{fig:GCET}, GCET first transforms the feature map of correspondences $F=\{f_1,\cdots,f_N\}$ into the graph network $\mathcal{G}_{i}=(\mathcal{V}_{i},\mathcal{E}_{i})$ where $\mathcal{G}_{i}$ denotes $i$-th correspondence, $\mathcal{V}_{i}=(v_{i}^{1}, \cdots, v_{i}^{k})$ contains $k$ neighbors and $\mathcal{E}_{i}=(e_{i}^{1}, \cdots, e_{i}^{k})$ indicates the relationships between $\mathcal{G}_{i}$ and its neighbors. Here, we describe $e_{i}^{j}$ as $[{f}_{i}\|{f}_{i}-f_{i}^{j}]$ where $[\cdot \| \cdot]$ means the concatenation operation. Then, the graph network is converted into two different types of graph context: credible graph context (CGC) and structure graph context (SGC) by maxpooling with MLPs and convolution with $p$ neighborhood segmentation \cite{zhao2021progressive} to gather diverse context information. This process can be expressed as follows:
\begin{align}
CGC&={MLPs}({Maxpooling}({MLPs}(\mathcal{E}_{i}))),\\
SGC&=Conv_{2}({Conv_{1}}(\mathcal{E}_{i})),
\end{align}
where $Conv_{1}$ and $Conv_{2}$ are convolutional operations with kernel sizes of $1\times p$ and $1\times \frac{k}{p}$, respectively.
\par Although CGC discards a majority of edge information, it remains the most credible neighbor relationships. Conversely, SGC, captures the most of structure information among nodes but is susceptible to interference from contaminated information. Next, in order to amplify the strengths of these graph contexts, we employe self-attention to recalibrate themselves and leverage cross-attention in parallel to uncover shared significant part. However, both self-attention and cross-attention demand substantial computation resources, especially when dealing with a large number of $N$. Therefore, before the recalibration of graph contexts, it is imperative to streamline graph contexts into $\{CGC^{\prime}, SGC^{\prime}\}$ through a clustering operation \cite{zhang2019learning} to compact vertices in a learnable manner. The detail operation can be formulated as follows:
{\small
\begin{align}
CGC^{\prime}, SGC^{\prime}&={Cluster}(CGC, SGC),\\
CGC^{\prime,e}&=({SA}(CGC^{\prime})\oplus CA(SGC^{\prime},CGC^{\prime})),\\
SGC^{\prime,e}&=({SA}(SGC^{\prime})\oplus CA(CGC^{\prime},SGC^{\prime})),
\end{align}
}where $CGC^{\prime,e}$ and $SGC^{\prime,e}$ denote enhanced graph contexts in a clustered state. $SA(\cdot)$ represents the self-attention and $CA(\cdot,\cdot)$ indicates the cross-attention where the query is derived from the preceding input and the key-value pairs source from the second input. Additionally, $\oplus$ is the attentional fusion operation \cite{dai2021attentional} which discriminately treats transitional graph contexts to generate the complete graph context with strong characteristics.
\par Finally, the enhanced graph contexts are recovered to the original sizes to keep the permutation invariance and go through another attentional fusion to combine respective highlighted advantages. The process can be described as:
{\small
\begin{align}
CGC^{e}, SGC^{e}&={Recover}(CGC^{\prime,e}, SGC^{\prime,e}),\\
GC^{e}&=(CGC^{e}\oplus SGC^{e}),
\end{align}
}where $GC^{e}$ is the output of GCET.
\subsection{Graph Context Guidance Transformer}

\par Global consensus is served as convictive evidence to assist in inlier/outlier discerning. Nevertheless, due to the substantial presence and random distribution of outliers, excavating global consensus among inliers is a highly challenging task. To extend the application of the enhanced graph context to the global realm, we design GCGT, to guide the inlier discrimination process by mining the consensus among inliers.
\par The detailed guidance process is shown in Fig.~\ref{fig:GCGT}. In specific, we first subject the enhanced graph context to a linear layer, assigning confidence scores to each node to generate a score table (ST). Based on ST, we proceed to sort confidence scores in descending order and sample vertices with higher confidence scores to form a set of candidates. Notably, before delving into the consensus guidance procedure, we expand the candidate set to enhance its expressive capacity and mitigate the potential disruption caused by hidden outliers. Simultaneously, we perform the cluster operation to the enhanced graph context before sampling phase, streamlining its representation and concurrently reducing the computational load during the guidance process. These preparations can be described as:
\begin{figure}[t]
\centering
\centerline{\includegraphics[width=0.48\textwidth]{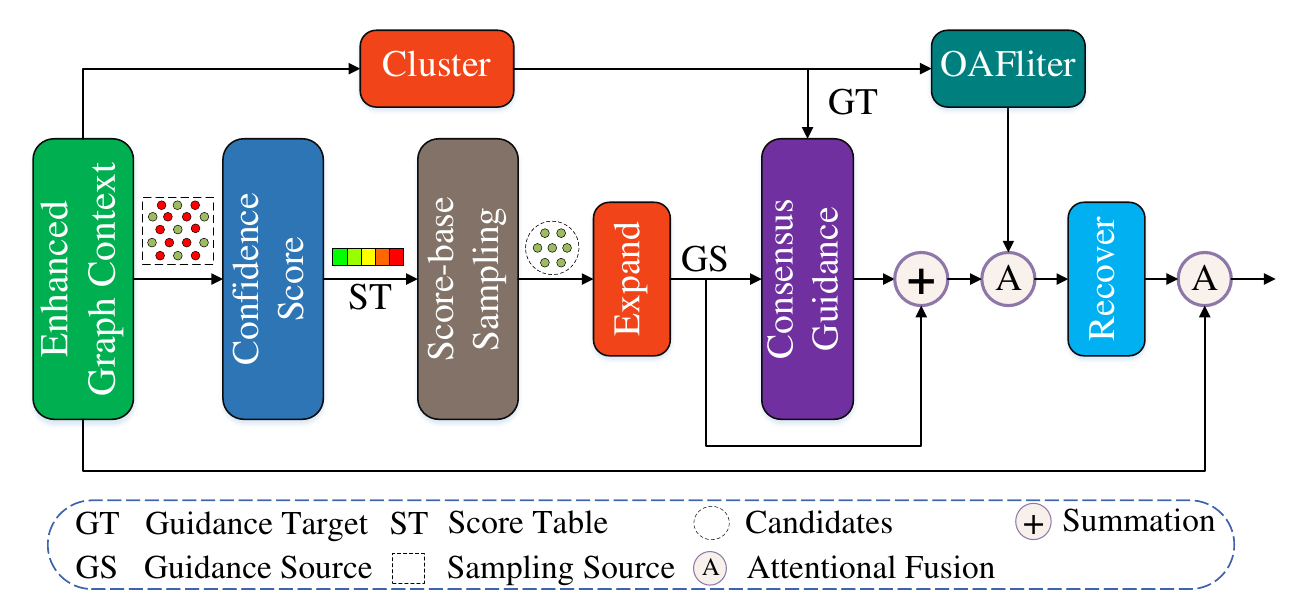}}
\caption{The detailed structure of our proposed graph context guidance transformer.}
\label{fig:GCGT}
\end{figure}
{
\begin{align}
ST&={Linear Layer}(GC^{e}),\\
GS&={Expand}({Sample}({Sort_{dec}}(ST,sr))),\\
GT&={Cluster}(GC^{e}),
\end{align}
}where $Sort_{dec}$ implies sorting targets in descending order, $sr$ indicates sampling rate, and GS and GT represents the guiding source and the guiding target, respectively. $Expand$ is the inverse operation of $Cluster$.
\par Next, we employ the vanilla Transformer to conduct the consensus guidance which seeks similarities between $GS$ and $GT$ to assign greater attention to inliers. Here, the query is a linear projection of $GS$ and the key-value pairs source from $GT$. To prevent from information loss during guidance procedure, we apply skip connection to GS. Besides, we also perform OAFliter \cite{zhang2019learning} to the clustered graph context, which captures spatial-wise dependencies complementing the output of Transformer. Finally, we recover the fusion results of the output of OAFilter and consensus guidance and further integrate $GC^{e}$ to harmonize the balance between local context and global consensus. These operations can be formulated as follows:
{\small
\begin{align}
GR&=(({TF}(GS,GT)+GS)\oplus OAFilter(GT)),\\
GC_{out}&=(Recover(GR)\oplus GC^{e}),
\end{align}
}where $GR$ means guiding results, $TF$ indicates the vanilla Transformer and $GC_{out}$ is the final graph context output by GCGT.
\subsection{Loss function}
\par Following \cite{yi2018learning}, we employ a hybrid loss function to optimize GCT-Net. The loss function is composed of two constituents:
\begin{align}
L&=L_{cls}\left(o_i , y_i\right)+\delta L_{reg}\left(\hat{E}, E\right)\label{eqn:loss},
\end{align}where $L_{cls}$ and $L_{reg}$ denotes the correspondence classification loss and the essential matrix regression loss, respectively. $\delta$ represents a parameter utilized to balance these two losses. $L_{cls}$ can be further formulated as:
\begin{align}
L_{cls}(o_i,y_i)=\sum_{i=1}^K\mathcal{H}(\eta_i\odot o_i,y_i),
\end{align}where $\mathcal{H}$ represents the binary cross entropy loss. $o_i$ signifies the logit values derived from $i$-th pruning module. $y_i$ denotes the ground-truth label set for the $i$-th pruning module where labels are ascertained by the threshold of $10^{-4}$. $\odot$ is the Hadamard product. The parameter $\eta_i$ is a dynamic temperature vector, strategically leveraged to mitigate the negative effects of label ambiguity \cite{zhao2021progressive}. $K$ indicates the count of correspondence pruning modules. $L_{reg}$ can be described as follows \cite{ranftl2018deep}:
{\scriptsize
\begin{align}
L_e(\hat{E},E)&=\frac{(p^{'T}\hat{E}p)^2}{\|Ep\|_{[1]}^2+\|Ep\|_{[2]}^2+\|E^Tp^{'}\|_{[1]}^2+\|E^Tp^{'}\|_{[2]}^2},
\end{align}
}where $p$ and $p^{'}$ denote the coordinate sets in image matching pairs. $q_{[i]}$ stands for the $i$-th element in vector $q$.
\section{Experiments}
\subsection{Evaluation Protocols}
\subsubsection{Datasets}
\par We conduct experiments on outdoor and indoor datasets ($i.e.$, YFCC$100$M and SUN$3$D) to demonstrate the outlier removal capability of GCT-Net. The YFCC$100$M dataset contains $100$ million publicly accessible travel images divided into $71$ sequences. The SUN$3$D dataset, comprising a substantial collection of RGBD images, has been categorized into $254$ sequences. As in \cite{zhang2019learning}, these sequences are further divided to generate a training set, a validation set, and a test set. Some images from the sequences being used as training set are retained to act as known scenario testing.
\subsubsection{Evaluation metrics}
\par We evaluate our proposed GCT-Net in terms of both inlier/outlier classification and relative pose estimation tasks. In the inlier/outlier classification task, the network is supposed to remove outliers and preserve as many inliers as possible. Therefore, \emph{Precision} ($P$), \emph{Recall} ($R$) and \emph{F-score} ($F$) are selected as our evaluation metrics. In relative pose estimation task, the mean average precision (mAP) is adopted as our criteria which measures the angular differences between estimated vectors and the ground truth ones with the perspective of both rotation and translation.

\subsection{Evaluation Protocols}
\par In the overall framework implementation of our network, following \cite{zhao2021progressive}, we utilize two consecutive pruning modules with a pruning rate of $0.5$ each to achieve progressive selection. SIFT is employed to generate an initial set of $N$ = $2000$ correspondences where the number of channel dimension $d$ is extended to $128$. In GCET, the neighbor number $k$ in KNN algorithm is set to $9$ for constructing the graph network. In GCGT, we configure the sampling rate $sr$ to be $0.2$. As for the common components in GCET and GCGT, the channel reduction ratio $r$ in attentional fusion \cite{dai2021attentional} and the head number $h$ in Transformer \cite{vaswani2017attention} is all configured to $4$. In alignment with configuration of \cite{zhang2019learning}, we utilize the Adam optimizer \cite{kingma2014adam}, to set the batch size to $32$ and maintain the learning rate of $10^{-3}$ to train our network. It's noteworthy that the training process spans a total of $500k$ epochs where for the initial $20$k epochs, $\delta$ in Eq.~\ref{eqn:loss} is set to 0, and for the remaining $480$k epochs, $\delta$ is fixed to $0.5$.
\begin{table}[t]
 \centering
 \scalebox{0.8}{
  \begin{tabular}{c c c c c c c }
   \toprule
   Datasets & \multicolumn{3}{c}{YFCC100M} & \multicolumn{3}{c}{SUN3D}\\
   \midrule
   \multirow{1}*{Methods} & $P$ (\%)  & $R$ (\%)   & $F$ (\%)  & $P$ (\%) & $R$ (\%) & $F$ (\%) \\
   \midrule
   RANSAC            & 43.51 & 50.68 & 46.82 & 44.89 & 48.68 & 46.71          \\
   LFGC              & 54.67 & 84.76 & 66.47 & 43.95 & 83.71 & 57.64           \\
   OANet++              & 55.78 & 85.93 & 67.65 & 46.15 & 84.36 & 59.66           \\
   ACNet          & 55.54 & 85.38 & 67.30 & 45.97 & 83.94 & 59.40         \\
   T-Net          & 58.21 & 86.38 & 69.55 & 47.27 & 84.16 & 60.54         \\
   CLNet            & 75.08 & 76.42 & 75.73 & 60.01 & 68.10 & 63.80         \\
   MS$^2$DG-Net     & 59.11 &  88.4 & 70.85 & 46.95  & 84.55 & 60.37         \\
   MSA            & 58.70 & 87.99 & 70.42 & 48.10 & 83.81 & 61.12         \\
   ConvMatch        & 58.77 & \textbf{89.39} & 70.92 & 47.54 & \textbf{85.38} & 61.07       \\
   Ours             & \textbf{77.00} & 79.02 & \textbf{78.00} &\textbf{61.12}  & 69.34 & \textbf{64.31}   \\
   \bottomrule
 \end{tabular}}
 \caption{Comparison of the Precision, Recall, and F-score between GCT-Net and other methods across the YFCC$100$M and SUN$3$D datasets for the correspondence classification task.}
 \label{tab:prfcompare}
\end{table}
\subsection{Correspondence Classification}
\par We perform a comprehensive comparison between GCT-Net and a selection of classic and cutting-edge works, spanning the traditional method \cite{fischler1981random} as well as learning-based methods \cite{yi2018learning,zhang2019learning,sun2020acne,zhong2021t,zhao2021progressive,dai2022ms2dg,zheng2022msa,zhang2023convmatch}. Here, we utilize a ratio test with a threshold of $0.8$ in RANSAC to proactively eliminate certain erroneous matches, preventing a sharp decline. 
\begin{figure}[t]
    \centering
    \begin{subfigure}[b]{0.24\linewidth}
        \includegraphics[width=0.98\linewidth]{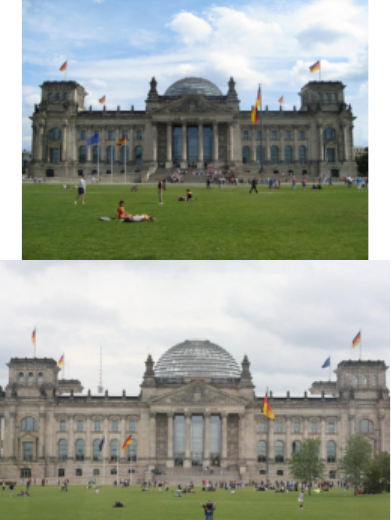}\\
        \includegraphics[width=0.98\linewidth]{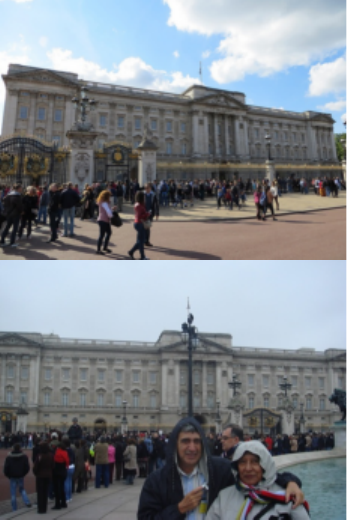}\\
        \includegraphics[width=0.98\linewidth]{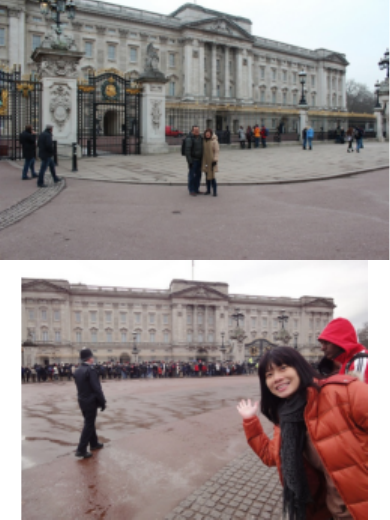}\\
        \includegraphics[width=0.98\linewidth]{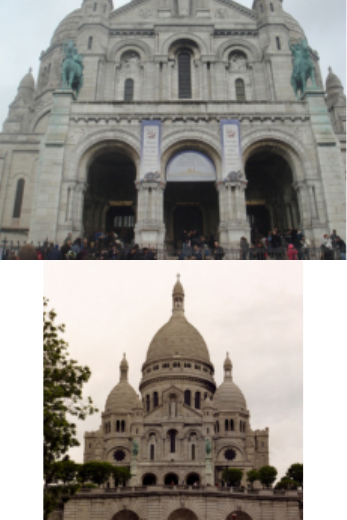}
        \caption{Original}
    \end{subfigure}
    \begin{subfigure}[b]{0.24\linewidth}
        \includegraphics[width=0.98\linewidth]{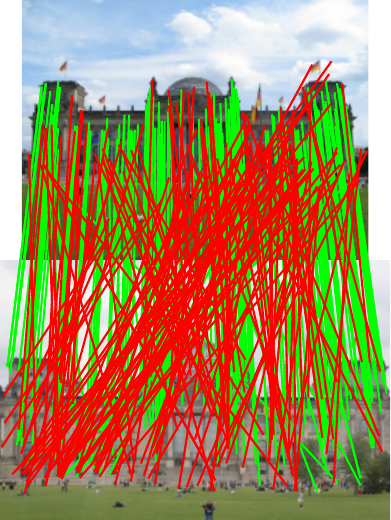}\\
        \includegraphics[width=0.98\linewidth]{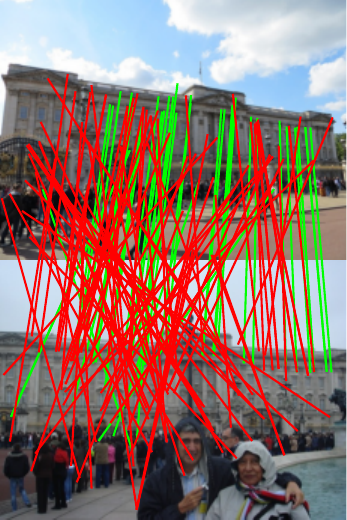}\\
        \includegraphics[width=0.98\linewidth]{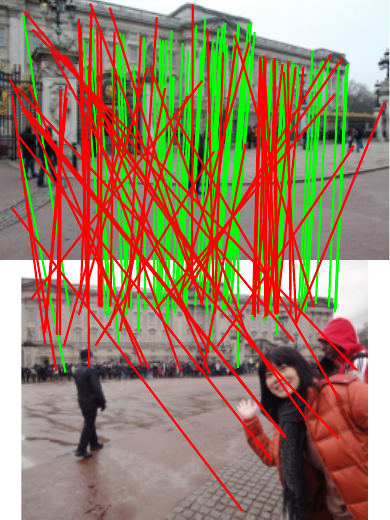}\\
        \includegraphics[width=0.98\linewidth]{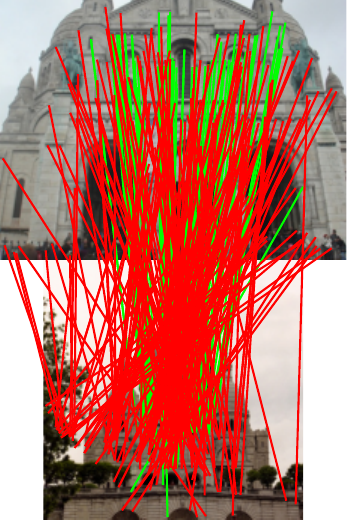}
        \caption{RANSAC}
    \end{subfigure}
    \begin{subfigure}[b]{0.24\linewidth}
        \includegraphics[width=0.98\linewidth]{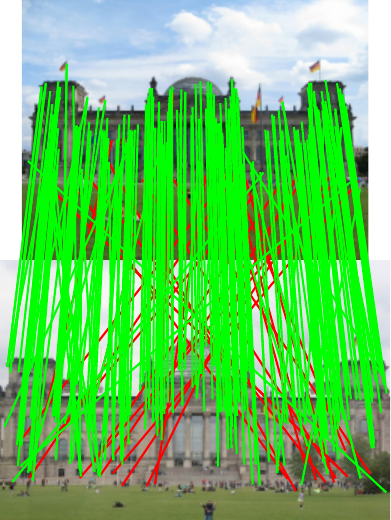}\\
        \includegraphics[width=0.98\linewidth]{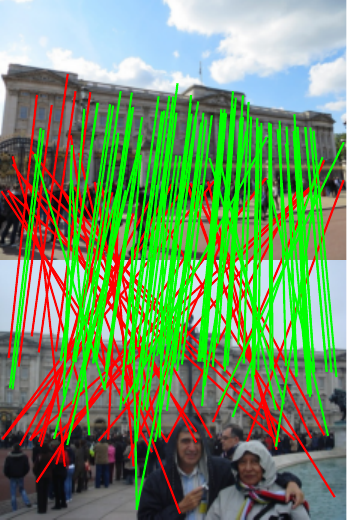}\\
        \includegraphics[width=0.98\linewidth]{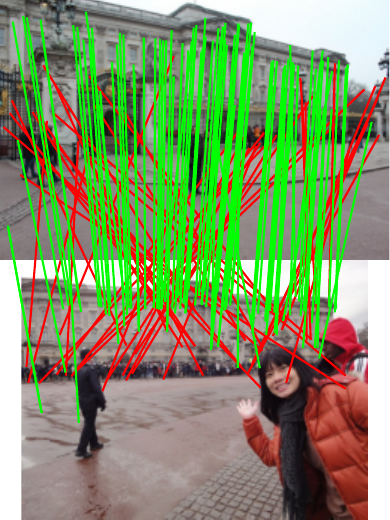}\\
        \includegraphics[width=0.98\linewidth]{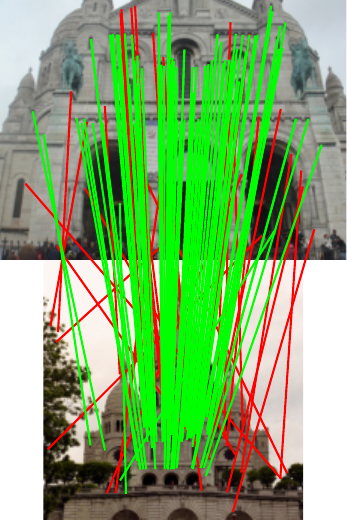}
        \caption{CLNet}
    \end{subfigure}
    \begin{subfigure}[b]{0.24\linewidth}
        \includegraphics[width=0.98\linewidth]{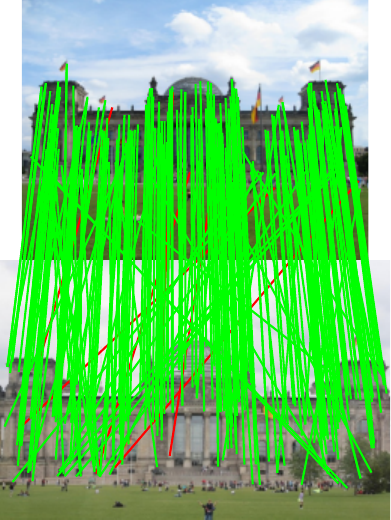}\\
        \includegraphics[width=0.98\linewidth]{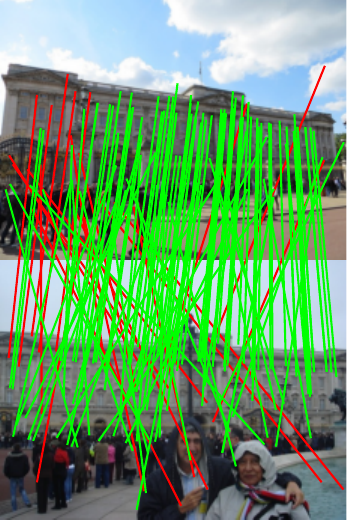}\\
        \includegraphics[width=0.98\linewidth]{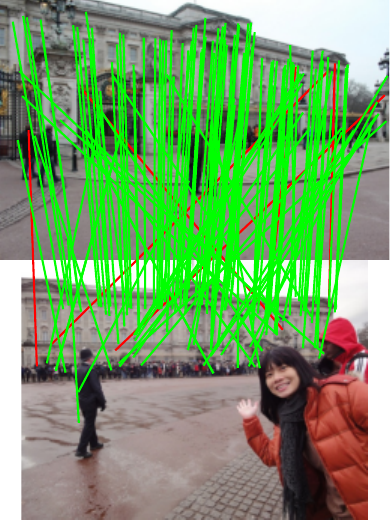}\\
        \includegraphics[width=0.98\linewidth]{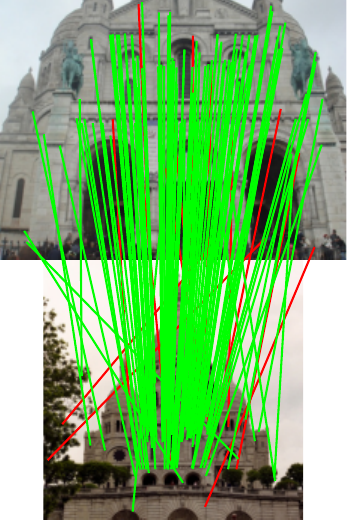}
        \caption{GCT-Net}
    \end{subfigure}
    \caption{Visualization results of correspondence pruning methods on outdoor datasets. From left to right: Original, RANSAC, CLNet, and GCT-Net. Green lines denote the preserved inliers and red lines indicate not removed outliers.}
    \label{fig:visual}
\end{figure}

\par Table \ref{tab:prfcompare} showcases the comparative results conducting the task of correspondence classification on YFCC$100$M and SUN$3$D. We can observe that our network achieves the best performance, except in terms of the Recall metric. The reason lies in our adoption of the progressive correspondence pruning strategy, which, while removing a mass of outliers, inevitably eliminates hidden inliers as well. Consequently, our method and CLNet exhibit significant improvement in the Precision metric, while the value of Recall is relatively lower compared to other methods. However, considering the overall metrics ($i.e.$ F-score), we still gain the optimal results which surpasses the second-best method by $2.27\%$ and $0.51\%$ on the YFCC$100$M and SUN$3$D datasets, respectively. Fig. \ref{fig:visual} displays the visualization results of classification, which further demonstrates the remarkable ability of our network in removing outliers.

\begin{table}[h]
\centering
 \scalebox{0.8}{
\begin{tabular}{cccccc}
\toprule
\centering
&\multirow{2}{*}{Matcher} & \multicolumn{2}{c}{Known} &\multicolumn{2}{c}{Unknown} \\ \cmidrule(r){3-4} \cmidrule(r){5-6}
&&5$ ^\circ$&20$ ^\circ$&5$ ^\circ$&20$ ^\circ$\\
\midrule
\multirow{9}{*}{SIFT}
                            & RANSAC &       5.81         &    16.88            &   9.07              &     22.92                   \\
                            & LFGC                 &    13.81            &       35.20         &       23.95          &     52.44                    \\
                            & OANet++                 &     32.57           &          56.89      &          38.95       &    66.85                       \\
                            & ACNet                 &     29.17          &      52.59          &      33.06         &      62.91                   \\
                            & T-Net                          &41.51 &65.12& 48.40&74.23        \\
                            & CLNet                          &38.27 &62.48& 50.80&75.76        \\
                            & MS$^2$DG-Net &38.36&64.04&49.13&76.04\\
                            & MSA                          &39.53 &61.75& 50.65&77.99        \\
                            & ConvMatch                          &43.48 &66.14& 54.62&77.24        \\
                            & Ours & $\textbf{49.05}$ & $\textbf{71.32}$ & $\textbf{63.80}$ & $\textbf{82.86}$\\
\midrule
\multirow{9}{*}{SuperPoint} & RANSAC   &      12.85          &           31.22     &     17.47            &     38.83                     \\
                            & LFGC                 &      12.18         &          34.75      &   24.25              &     52.70                      \\
                            & OA-Net++                 &    29.52            &    53.76            &     35.27            &      66.81                   \\
                            & ACNet                &   26.72             &      49.29          &   32.98              &     62.68                     \\
                            & T-Net                          &34,97 &57.50& 40.65&70.36        \\
                            & CLNet                          &27.56 &50.82& 39.19&67.37        \\
                            & MS$^2$DG-Net                  &31.15&55.16&39.19&70.36 \\
                            & MSA                          &30.63 &53.74& 38.53&68.56\\
                            & ConvMatch                  &$\textbf{38.34}$&60.25&$\textbf{48.80}$&74.59 \\
                            & Ours &37.86 & $\textbf{60.55}$ & 47.17 & $\textbf{74.76}$\\
                            \bottomrule	
\end{tabular}}
\caption{Comparison results about GCT-Net and alternative methods on known and unknown scenes in YFCC$100$M under the mAP5$^{\circ}$ and mAP20$^{\circ}$ metrics for the relative pose estimation task.}
\label{tab:regress}
\end{table}
\par After correspondence classification, inliers are assigned to weights to execute the relative pose estimation task. The concerning experimental results are shown in Table~\ref{tab:regress}. Here, we also evaluate the compatibility of various feature matching methods with different feature extraction approaches. In contrast to the hand-crafted method SIFT \cite{fischler1981random}, we employ a learning-based feature extraction approach, SuperPoint \cite{detone2018superpoint} for testing. In experiments, we select mAP$5^\circ$ and mAP$20^\circ$ to comprehensively evaluate the performance of these methods under high-tolerance and low-tolerance scenarios. Besides, to assess the generalization capability of models, we conduct experiments in both known and unknown scenes.

\par From Table~\ref{tab:regress}, it is apparent that GCT-Net outperforms all configurations under the SIFT-based condition. When compared to CLNet, which also employs the progressive pruning framework, our network demonstrates a significant lead in unknown scenes, surpassing CLNet by $13\%$ in mAP$5^\circ$ and $7.1\%$ in mAP$20^\circ$. We also achieve $9.18\%$ and $6.98\%$ improvements compared to ConvMatch on unknown and known scenes under mAP$5^\circ$. However, when adopting SuperPoint as the feature extraction method, our network only slightly surpasses ConvMatch in mAP$20^\circ$, while trailing behind ConvMatch in mAP$5^\circ$. This discrepancy might be attributed to SuperPoint generating lots of high-quality correspondences at the beginning. For our network, pruning such high-quality correspondences could result in the loss of crucial information and thus cause a decrease in estimation accuracy. In contrast, ConvMatch can leverage convolutions to capture additional information effectively.
\subsection{Ablation Studies}
\begin{table}[t]
 \centering
  \scalebox{0.75}{
\begin{tabular}{cccc|ccc}

\toprule

\makecell[c]{IPS} &\makecell[c]{GCET}& \makecell[c] {GCGT-P}  &\makecell[c]{GCGT-W} &\makecell[c]{ mAP5$^{\circ}$} &\makecell[c]{mAP20$^{\circ}$}        \\
\midrule

\multirow{1}*{$\surd$}   &&&                                                    &46.63&71.06\\
\multirow{1}*{$\surd$}   &$\surd$&&                                             &60.60&81.33\\
\multirow{1}*{$\surd$}   &&$\surd$&                                             &57.01&78.70\\
\multirow{1}*{$\surd$}   &&&$\surd$                                         &58.30&79.54\\
\multirow{1}*{$\surd$}   &$\surd$&&$\surd$                             &63.80&82.86\\
\bottomrule
\end{tabular}}
\caption{Overall ablation studies on the YFCC100M dataset.}
\label{tab:overall_ablation}
\end{table}
\par We perform ablation experiments on GCT-Net to demonstrate the effectiveness of individual components. Table~\ref{tab:overall_ablation} displays the experimental results about the network integration with various modules. IPS indicates that the network composed of only ResNet blocks adopts the iterative pruning strategy. GCET represents the application of Graph Context Enhance Transformer. GCGT-P refers the partial Graph Context Guidance Transformer, where we isolate the injection of OAFilter to assess the effectiveness of the process of sampling to consensus guidance. GCGT-W signifies the whole Graph Context Guidance Transformer.
\par From Table~\ref{tab:overall_ablation}, it is evident that the integration of every component causes a favorable impact on the network performance, compared to single utilization of IPS. In specific, the second row of table which incorporates GCET into IPS obatins $13.97\%$ and $10.27\%$ improvments under mAP5$^{\circ}$ and mAP20$^{\circ}$. This demonstrates that the significance of generating graph contexts and effectively leveraging them. The third row ($i.e.$ IPS $+$ GCGT-P) validates the effectiveness of the sampling strategy and consensus guidance, which gains a $11.67\%$ improvement under mAP5$^{\circ}$. Compare to partial GCGT, utilizing complete GCGT (the fourth row) results in improvements of $1.3\%$ and $0.84\%$ under mAP$5^{\circ}$ and $20^{\circ}$. This highlights the injection of OAFilter, which enhances the output of Transformer in a complementary manner. By combining GCET and GCGT, the network achieves the optimal performance.

%

\begin{figure}[t]
    \centering
    \begin{subfigure}[b]{0.49\linewidth}
        \includegraphics[width=\linewidth]{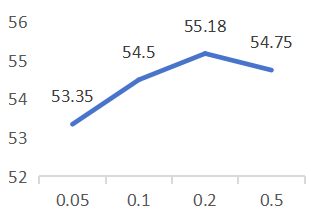}\\
        \caption{mAP$5^{\circ}$}
    \end{subfigure}
    \begin{subfigure}[b]{0.49\linewidth}
        \includegraphics[width=\linewidth]{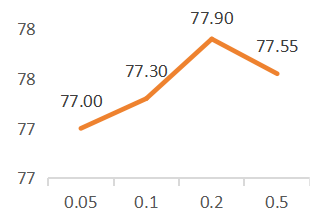}\\
        \caption{mAP$20^{\circ}$}
    \end{subfigure}
    \caption{Ablation studies about the selection of different sampling rates. The $x$-$axis$ and $y$-$axis$ represent the sampling rate and the evaluation metric, respectively. }
    \label{fig:sp}
\end{figure}
\par We also perform the ablation studies about different sampling rates. When the sampling rate is excessively large, it can lead to substantial computational load. Therefore, during the implementation of experiments, we maintain the sampling rate below $0.5$. As shown in Fig.~\ref{fig:sp}, opting for a low sampling rate ($i.e.$, $0.05$) can limit the expressive capacity of network, whereas selecting an excessively high sampling rate ($i.e.$, $0.5$) can make it susceptible to disruption by outlier information. Therefore, it's necessary to choose an appropriate sampling rate ($i.e.$, $0.2$) to strike a balance between the two aspects.

\section{Conclusion}
\par In this paper, we propose the effective Graph Context Transformation Network (GCT-Net) for progressive correspondence pruning. The graph network is served as an effective carrier of local context information. Therefore, we propose the Graph Context Enhance Transformer to convert the graph network into multi-branch graph contexts and enhance the individual characteristic and shared significant information of graph contexts. This allows the advantages of different graph contexts to be effectively combined and fully utilized. For extending the enhanced graph context to the global domain, we further design the Graph Context Guidance Transformer. This module adopts a score-based sampling strategy to select candidates as the guiding source and regards the unsampled vertices as the guiding target for the execution of consensus guidance which seeks the hidden inliers by consensus similarities. Numerous experiments conducted on tasks related to correspondence classification and relative pose estimation demonstrate the superior ability of GCT-Net, surpassing the performance of state-of-the-art methods.
 \section*{Acknowledgment}%
{\small This work was supported by the National Natural Science Foundation of China under Grants 62072223, 62125201 and 62020106007.
\bibliography{mybib}}
\end{document}